\documentclass[runningheads]{llncs}
\usepackage{graphicx}
\usepackage{amssymb}
\usepackage{amsmath}
\usepackage{makecell}
\usepackage{hyperref}
\usepackage{xcolor}
\usepackage{cite}
\begin{document}
\author{Jaden Mu \\ jaden.mu@gmail.com
}
\institute{East Chapel Hill High School}

\title{A Real-Time Defense Against Object Vanishing Adversarial Patch Attacks for Object Detection in Autonomous Vehicles}
\titlerunning{
ADAV: Adversarial Defense for Autonomous Vehicles
}
\maketitle              
\begin{abstract}
Autonomous vehicles (AVs) increasingly use DNN-based object detection models in vision-based perception.  Correct detection and classification of obstacles is critical to ensure safe, trustworthy driving decisions.  Adversarial patches aim to fool a DNN with intentionally generated patterns concentrated in a localized region of an image.  In particular, object vanishing patch attacks can cause object detection models to fail to detect most or all objects in a scene, posing a significant practical threat to AVs.

This work proposes ADAV (Adversarial Defense for Autonomous Vehicles), a novel defense methodology against object vanishing patch attacks specifically designed for autonomous vehicles.  Unlike existing defense methods which have high latency or are designed for static images, ADAV runs in real-time and leverages contextual information from prior frames in an AV's video feed.  ADAV checks if the object detector's output for the target frame is temporally consistent with the output from a previous reference frame to detect the presence of a patch.  If the presence of a patch is detected, ADAV uses gradient-based attribution to localize adversarial pixels that break temporal consistency.  This two stage procedure allows ADAV to efficiently process clean inputs, and both stages are optimized to be low latency.  ADAV is evaluated using real-world driving data from the Berkeley Deep Drive BDD100K dataset, and demonstrates high adversarial and clean performance.

\end{abstract}
\section{Introduction}
Deep neural network (DNN)-based object detection models are increasingly used in autonomous vehicles (AVs).  For example, automotive company Tesla has already deployed a YOLO object detection model in its Full Self Driving software \cite{eduonix2022}.  

\begin{figure}
\includegraphics[width=\textwidth]{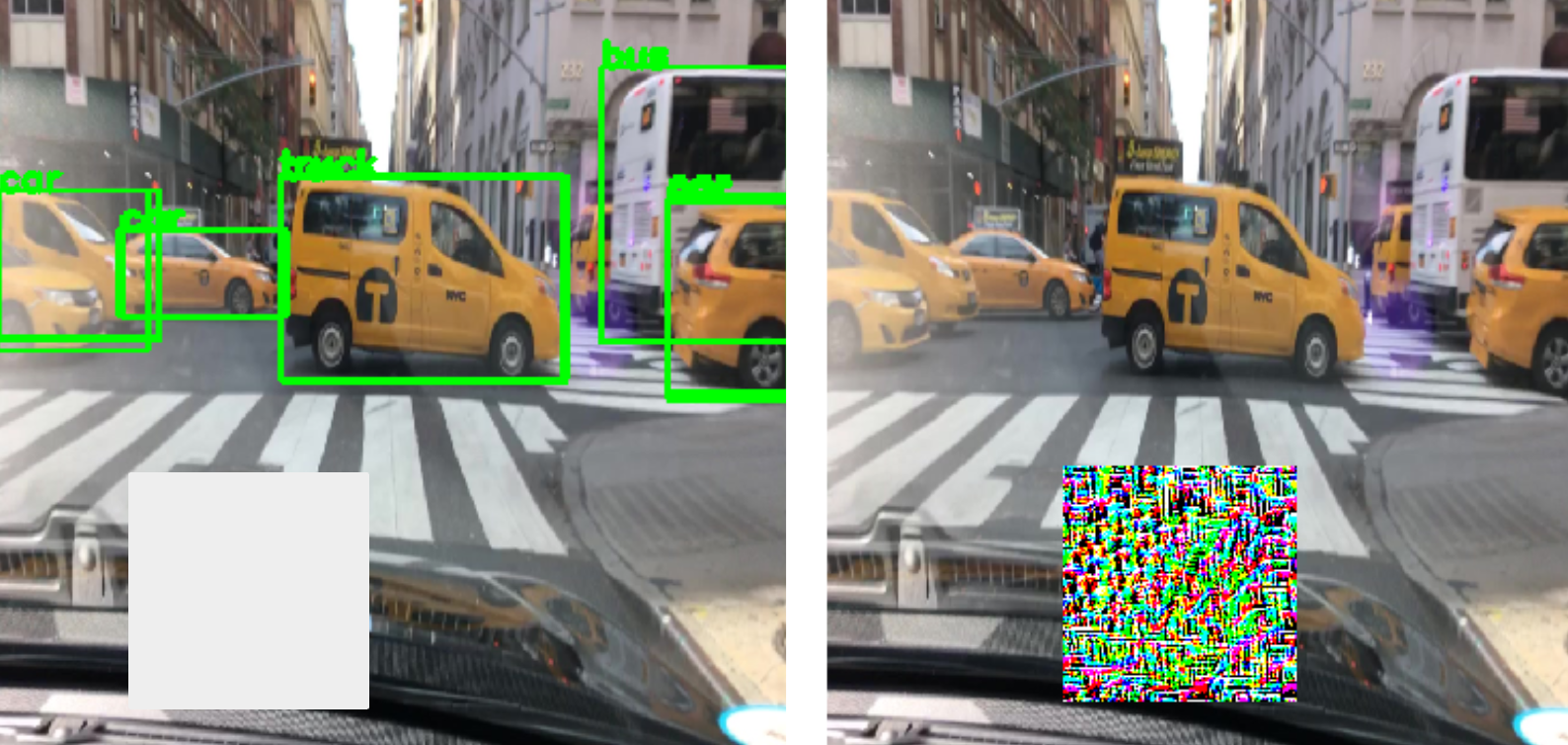}
\caption{The Object Vanishing Adversarial Patch}\label{fig1}
\end{figure}

Unfortunately, object detection models have been proven to be vulnerable to adversarial attacks \cite{liu2019}.  Adversarial patch attacks, which are spatially-constrained adversarial patterns designed to fool DNNs, are especially threatening due to their practicality, as they are robust to real-world transformations such as changes in position, scale, rotation, and perspective \cite{lee2019}.  Specifically, adversarial patch attacks that are trained on an object vanishing objective can lower an object detector's confidence on each prediction sufficiently to suppress all detections in an input image.  A patch attack can be conducted by overlaying a patch on an image.  In a real world autonomous driving context, this could be done by printing a physical patch and placing it on objects such as road signs \cite{tsuruoka2024}. As illustrated in Fig~\ref{fig1}, the object vanishing adversarial patch in the second image causes the object detector to fail to detect all objects in the input image. Note that this effect is not caused by overlapping, as the effect does not occur with the white (benign) patch in the first image.

Object detection in AVs is performed on video data, since AVs must constantly process a stream of sequential image data to make real-time decisions.  Object vanishing patch attacks can be applied to videos by applying an adversarial patch to each individual frame.  However, existing defenses have focused on defending object detection models on singular images.  This is reflected by the frequent use of datasets containing unrelated images for training and evaluation, such as ImageNet or COCO \cite{lee2019}, \cite{xiang21c}.  Therefore, existing defenses don't take advantage of the contextual information previous frames in a video can provide to the frame being processed.  Furthermore, existing defenses often don't prioritize latency, and are consequently unable process a video in real-time. 

An effective defense should \textbf{1)} Run in real-time, \textbf{2)} Recover a non-adversarial input from an adversarially attacked scene or mitigate the effects of the patch on the model's output, and \textbf{3)} Maintain good clean performance (the model paired with the defense should not perform significantly worse than the model on its own when there's no adversarial patch in frame).

ADAV improves on existing defenses by taking into account information provided by previous frames in an AV's video feed through \emph{temporal consistency}.  An object detector should detect the same objects in similar locations in frames close to each other temporally.  For example, a car detected in front of an AV should still be detected in a similar location half a second later - the typical environment an AV is in is consistent.  However, if an adversarial patch enters the field-of-view of the AV between the two frames and causes the model to not detect any objects, the model's output will be changed significantly, breaking temporal consistency.  ADAV measures the similarity in output between the target frame and a prior reference frame to determine if temporal consistency is broken, and therefore detect the presence of an adversarial patch.  If a patch is detected, ADAV then uses gradient-based pixel attribution to find the pixels that had the greatest contribution to changing the outputs between frames, which are likely adversarial pixels.  This allows ADAV to eliminate unnecessary computations and maintain clean performance by skipping the second stage if temporal consistency is not broken.  Additionally, parameters determining the threshold at which an adversarial patch is detected and localized are carefully tuned to balance the tradeoff between adversarial performance and clean performance.  We evaluate ADAV by training adversarial patches to apply to the BDD100K dataset \cite{yu2020}, and conduct several experiments to demonstrate strong adversarial and clean performance.

\section{Background}
\subsection{Gradient-Based Attribution}
Saliency Maps with vanilla gradients compute the gradient of the class score of interest with respect to the pixels in the input image \cite{simonyan2014}.  This creates a fine-grained map of the pixels in the input image which have the greatest impact on increasing the classification score.  Gradient based attribution methods/saliency maps have been widely used as an AI explainability method to visualize the regions of an image a model assigns the greatest importance.

Importantly for this work, saliency maps can also be computed for any differentiable function applied to the model's output, not just a class activation score, by extending the gradient calculation.

Guided backpropagation was proposed by \cite{springenberg2015} as a way to create cleaner saliency maps.  Guided backpropagation introduced a modification to vanilla gradients by only backpropagating positive gradients through ReLU activation functions, with the intuition that only focusing on gradients that increase the output would ensure that salient features don't get canceled out, and to produce a less noisy visualization.

\subsection{Existing Defenses}
Several defenses have been proposed to defend image classification models against adversarial patch attacks.  Notably, Local Gradient Smoothing (LGS) achieved good results in defending image classification by performing gradient smoothing on an image based on the observation that adversarial patches concentrate high frequency noise \cite{naseer2019}.  JPEG Compression was also proposed to defend against adversarial patch attacks, with the intuition that compression algorithms preserve important details while corrupting highly noisy patch reasons \cite{ferrari2023}.  Both LGS and JPEG Compression are model-agnostic, as they only preprocess the image.  Therefore, these defenses can also be applied in an object detection context.

Stronger defenses such as PatchGuard exist but are expensive to compute, with PatchGuard requiring several rounds of inference on smaller regions of an image to produce one certified output.

Fewer defenses exist for object detection.  DetectorGuard \cite{xiang2021} proposed a certified defense that issues an alert when a patch attack is detected.  DetectorGuard uses a robust image classifier that relies on a features extracted with a small receptive field, which ensures that patches can only have a small localized impact, to determine the approximate location of any objects in an image.  It then issues an alert if there is a mismatch between the image classifier and object detector.  However, DetectorGuard is a human-in-the-loop defense, and is incapable of addressing the attack beyond detecting the presence of a patch - a human must intervene if an alert is issued.  Universal Defense Frames \cite{yu2022} proposed preprocessing an input image by surrounding it with a "frame" of precomputed pixels trained to negate the effects of an adversarial patch, and requires almost no inference-time computations.

\section{Method}
\begin{figure}[t!]
\includegraphics[width=\textwidth]{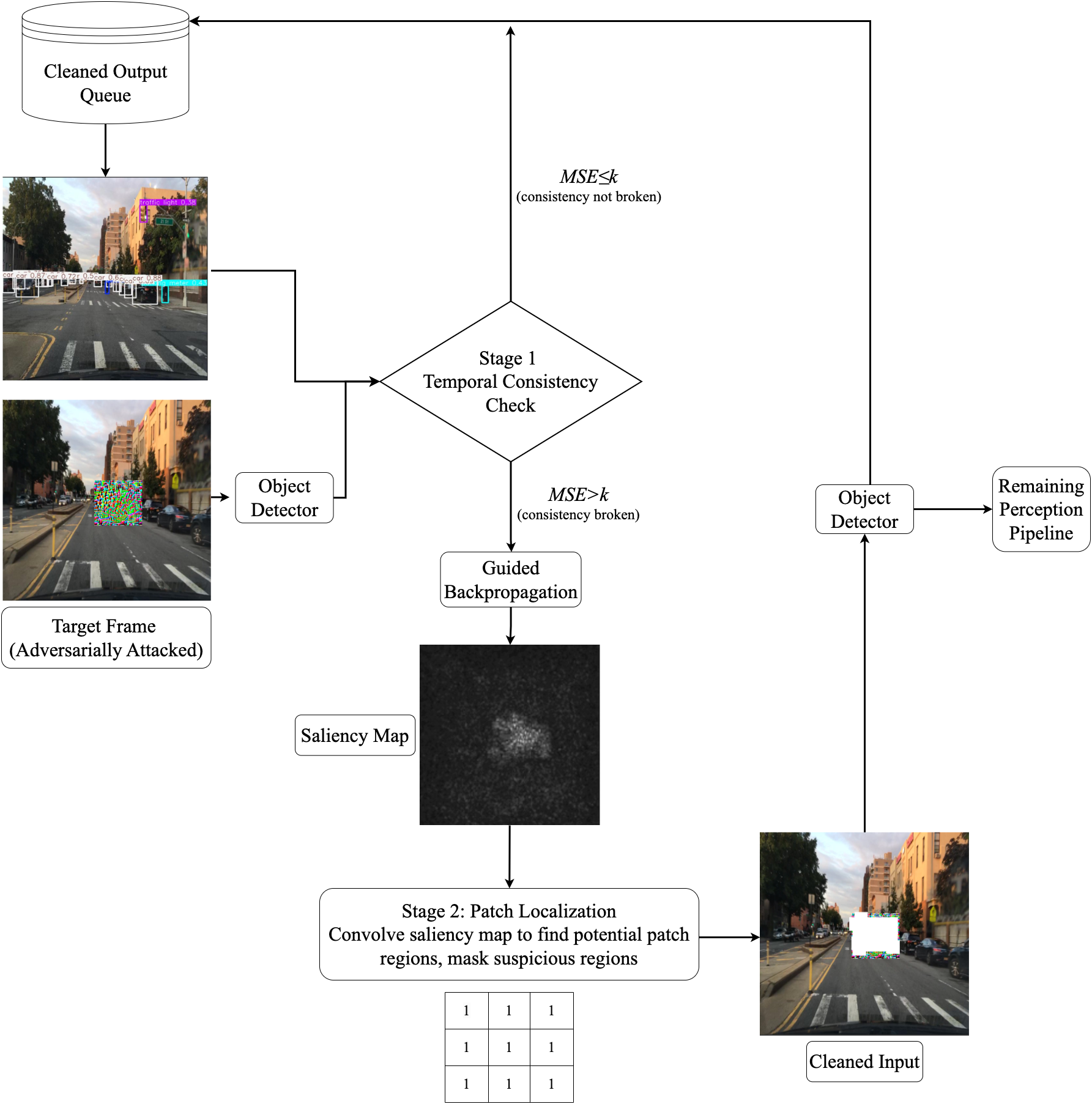}\caption{ADAV Methodology}\label{fig2}
\end{figure}

ADAV uses a two-stage process to defend against adversarial patches - it first checks if there is a patch anywhere in the image.  If the presence of a patch is detected, ADAV then localizes and masks the patch to produce a clean frame.  This process is illustrated in Fig~\ref{fig2}.

\subsection{Object Detection}
This work uses YOLOv5s \cite{jocher2020} as the base object detection model, although ADAV can be applied to other models.  YOLOv5 is a one-stage object detector that outputs 25200 potential bounding boxes.  The bounding boxes with the highest confidences are then returned with Non-Max Suppression.  Importantly, YOLOv5's one-stage architecture is fully differentiable.

\newpage
\subsection{Patch Detection}
Intuitively, the same objects should be detected in similar locations between two frames close to each other temporally.  This work refers to this concept as temporal consistency.  When an object vanishing adversarial patch enters into the field of view of an AV between two temporally close frames, temporal consistency is broken, since the adversarial patch will suppress several detections in the model's output.  This is leveraged to detect the presence of an adversarial patch.  ADAV checks if temporal consistency is broken between the frame being processed (the target frame) and a temporally close reference frame.  ADAV uses the frame 0.5 seconds before the target frame as the reference frame, which is assumed to be clean.  To quantify temporal consistency, the Mean-Squared Error (MSE) is computed between the model's output for the target frame and the reference frame from half a second prior.  The MSE of the outputs of YOLOv5s is substantially higher between 2 frames from 0.5 seconds apart, with one frame having an adversarial patch (\begin{math}\mu=46.9, \sigma=4.7\end{math}), than between 2 clean frames from 0.5 seconds apart (\begin{math}\mu=25.5, \sigma=6.2\end{math}).

Temporal Consistency is considered to be broken if the MSE in the model's output between 2 frames is greater than some threshold $k$, which is an empirically tuned parameter.  By checking if temporal consistency is broken before performing the relatively computationally expensive patch localization and masking, ADAV can improve latency on clean images by skipping the patch localization step.

To ensure the reference frame is clean, ADAV stores all outputs of clean frames in a queue.  The queue is initially populated with the outputs for the 0.5 seconds from the video feed, which is assumed to be clean (or cleaned with a methodology such as LGS).  Given the first 0.5 seconds as reference, ADAV then takes over, appending to the cleaned output queue with the output for each frame it cleans. 

\subsection{Patch Localization and Masking}
If a patch is detected (temporal consistency is broken), ADAV will localize and mask the patch to recover a clean input.  To localize the adversarial patch, this work uses guided backpropagation to create a saliency map.  However, instead of taking the gradient of a class activation, ADAV takes the gradient of the MSE between the outputs of the model for the target frame and the reference frame with respect to the input image.  Because adversarial patches break temporal consistency and therefore significantly increase the MSE, the saliency map should primarily flag pixels in the adversarial patches.  

Potential patch regions are extracted by downsampling the saliency map using strided convolutions.  Specifically, a 20x20 box filter kernel with a stride of 5 is used to sum the gradients in each potential patch region.  We refer to this sum of gradients as the "suspicion score" of a region.  Because adversarial patches produce very dense clusters on the saliency map, all suspicion scores below some threshold \textit n are ignored.  All remaining potential patch regions are assumed to be from the adversarial patch, so the corresponding pixels in the image are masked out.  

A small potential patch region size of 20x20 is chosen so that ADAV can neutralize patches with highly irregular, non-rectangular shapes by approximating the irregular shape with several small 20x20 regions.  Additionally, a small potential patch region helps minimize information loss if the region is falsely determined to be adversarial.

The threshold \textit n is determined dynamically for each image, since different images may have different magnitudes of gradients in clean regions.  Given a median suspicion score $\tilde{x}$ and interquartile range $Q$ of an image, $n=\tilde{x}+\lambda Q$, where $\lambda$ is an empirically tuned constant.  

Intuitively, this threshold detects outlier suspicion scores.  This should avoid false positives from sudden non-adversarial road condition changes in the 0.5 second interval (e.g. lighting changes), because the benign image will have no significant outliers.

To retrieve a cleaned input image, all pixels corresponding to a potential patch region with a suspicion score above $n$ are replaced with a neutral color which has no adversarial properties.  The model then generates a clean output on the cleaned input image.

\subsection{Cleaned Output Queue}
ADAV must have access to clean outputs to use as the reference.  This is done by storing clean outputs in the Cleaned Output Queue, which is initialized from the first 0.5 seconds of frames.  Then, if no patch is detected, the output of the YOLO model is directly added to the queue to use as a reference.  If a patch is detected, ADAV cleans the input, reruns the YOLO model, and adds the cleaned output to the queue, ensuring that the queue is always populated with 0.5 seconds of clean outputs.  Importantly, the outputs of the YOLO model are stored directly in the queue, meaning that no inference has to be repeated when checking for temporal consistency.

\subsection{Parameter Tuning}
The threshold for determining the presence of a patch $k$ and for filtering potential patch regions $\lambda$ must be tuned to be sensitive to the presence of an adversarial patch while remaining high enough to maintain clean performance.  Therefore, tuning these parameters must take into account the tradeoff between clean performance and adversarial performance.  Additionally, $k$ and $\lambda$ must be simultaneously optimized, since a lower $k$ might require a higher $\lambda$ to avoid false positives and vice versa.  We weight adversarial performance equally with clean performance.  The performance is measured by Mean Average Precision (mAP).  Letting $\pi_{k, \lambda}$ be a model defended with ADAV with parameters set to $k$ and $\lambda$, \[a=\max_{k, \lambda} [mAP(\pi_{k, \lambda}(x_{adv}))]\]  \[b=\max_{k, \lambda}[mAP(\pi_{k, \lambda}(x_{clean}))]\] 
\begin{equation*}
\begin{split}
k_{optimal}, \lambda_{optimal}=\arg\min_{k, \lambda}[(a - mAP(\pi_{k, \lambda}(x_{adv})) + (b - mAP(\pi_{k, \lambda}(x_{clean}))]
\end{split}
\end{equation*}
We find $k_{optimal}$ and $\lambda_{optimal}$ with a grid search.   

\begin{figure}
\begin{center}
\includegraphics[scale = 0.3]{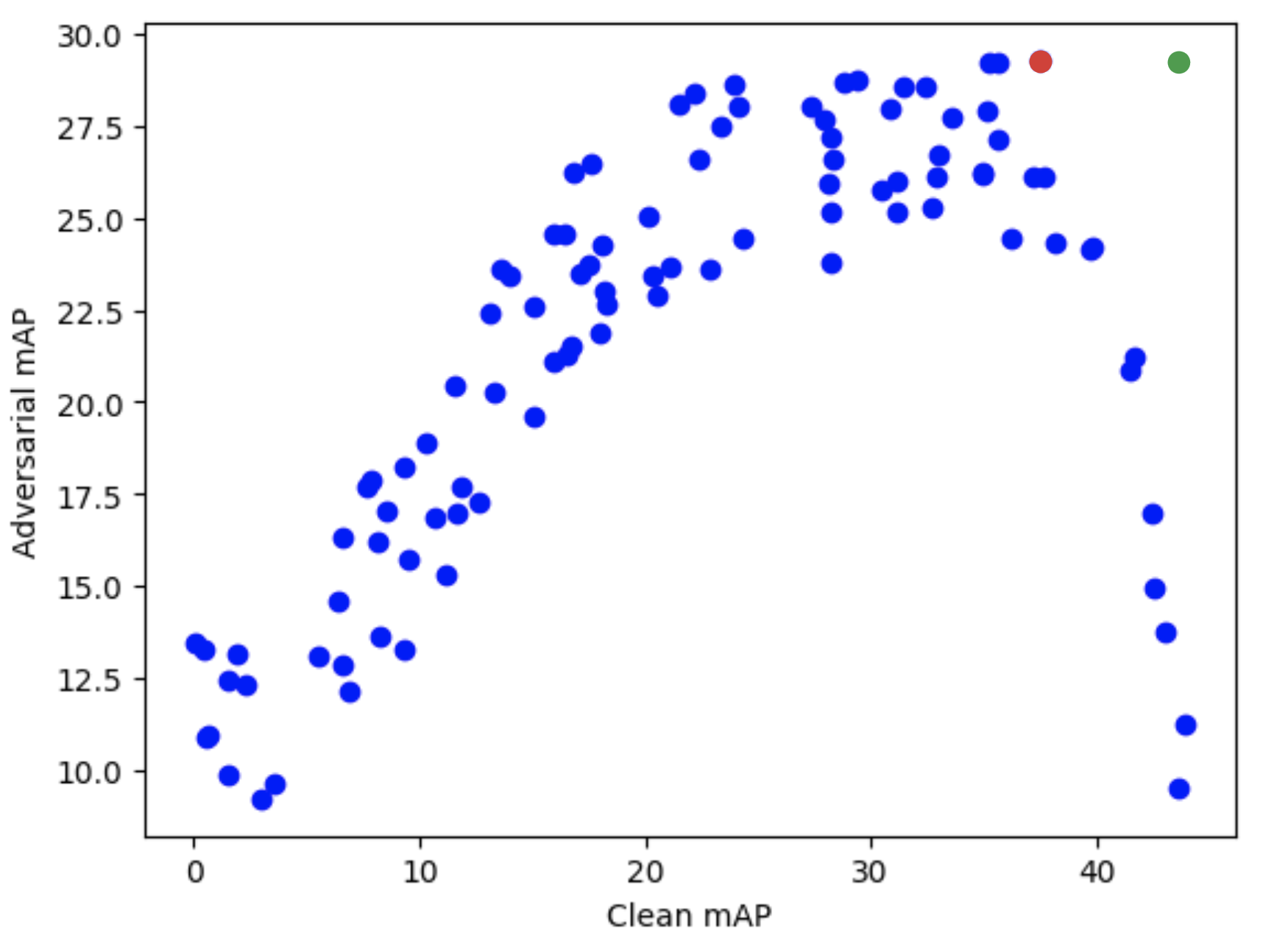}
\caption{Adversarial vs Clean Performance}\label{fig3}
\end{center}
\end{figure}

In Fig~\ref{fig3}, the blue and red points represent adversarial and clean performance of $\pi_{k, \lambda}$ for 100 samplings of $k$ and $\lambda$.  The red point represents the performance of $\pi_{k_{optimal}, \lambda_{optimal}}$, and the green point represents $(a, b)$.

\section{Experiments}
\subsection{Dataset}
This work uses the BDD100K dataset \cite{yu2020} for training YOLOv5s, generating adversarial patches, and for evaluation.  The BDD100K dataset is composed of 100,000 ~40 second long videos recorded at 30 frames per second (FPS) from vehicle dashcams.  BDD100K is a diverse dataset containing several vehicle and object types from multiple cities in different weather conditions and times of day.


\subsection{Object Detector Training}
To better simulate object detection in a self-driving context, YOLOv5s was trained from scratch on the 70000 image train split of BDD100K for 300 epochs, achieving 0.47 mAP. 


\subsection{Attack Formulation}
This work generates adversarial patches using the methodology proposed by \cite{pavlitskaya2022}, which found that finding a patch that maximizes YOLOv5s' confidence loss produces an effective object vanishing attack by lowering the confidence for each predicted bounding box below the confidence threshold for detection.  Specifically, a patch \textit P is generated by solving the following optimization problem:
\[\arg\max_P \mathbb{E}_{x \sim X,t \sim T}[\mathcal{L}(A(x,t,P), \hat{y})]\]
where \begin{math}\mathcal{L}\end{math} represents the YOLO confidence loss function, $t$ represents a transformation sampled from a distribution of transformations $T$ (changes in position, changes in scale), \textit A represents a patch applier function that applies patch \textit P at a position and scale determined by $t$ on $x$, and \begin{math}\hat{y}\end{math} represents the ground truth label for \textit x.

Training was done using projected gradient descent and the Adam optimizer with the learning rate set to \begin{math}0.1\end{math}.  A random image and transformation is sampled each training step.  This ensures that the patch is universal (able to attack all images) and robust to real-world transformations.  

A 200$\times$200 pixel patch taking 9.8\% of the input image was the patch size trained.  However, the actual patch size and position changes when the patch is applied to videos for evaluation.

\begin{figure}
\centering
\includegraphics[scale = 0.2]{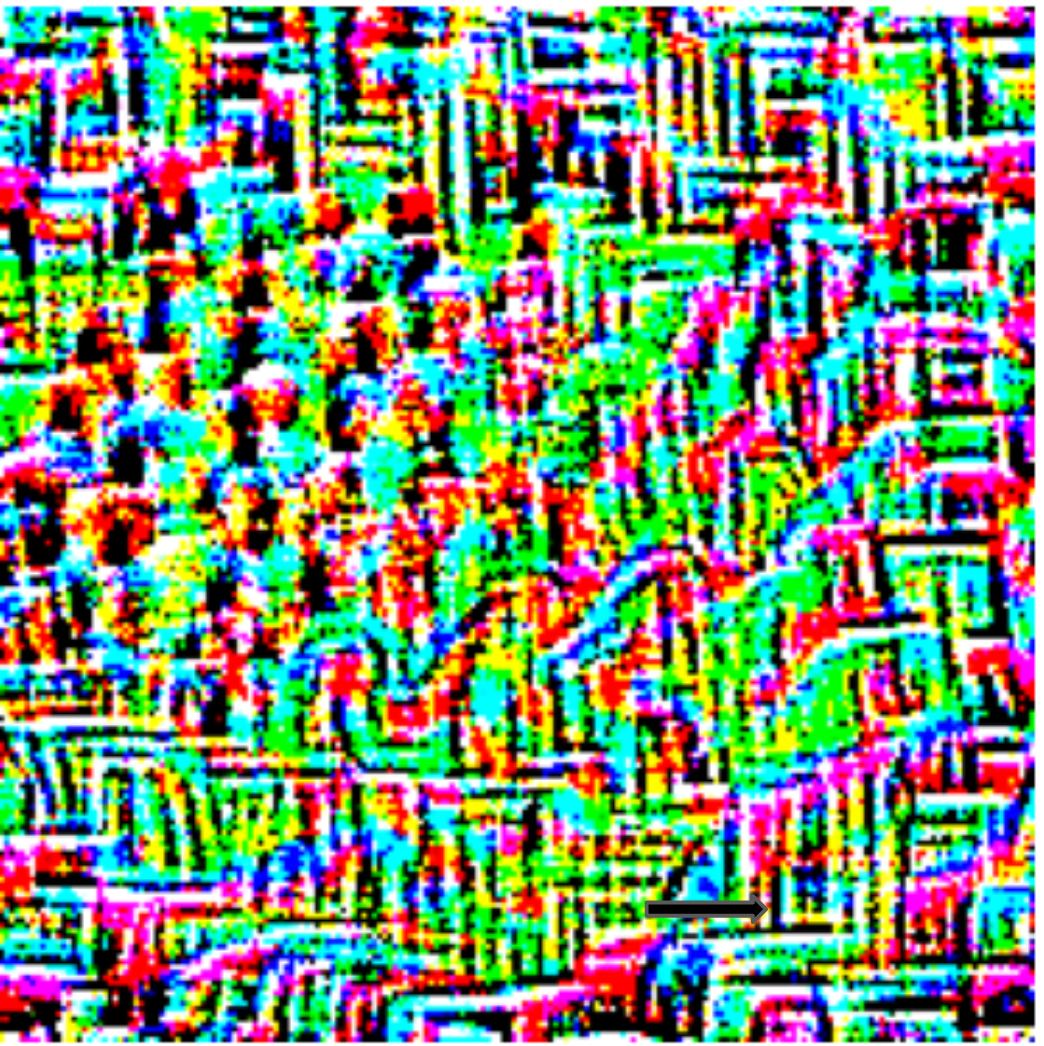}
\caption{Adversarial Patch}\label{fig6}
\end{figure}

\subsection{Synthetic Adversarial Video Creation}
In a realistic attack scenario, adversarial patches are placed on objects in motion relative to the AV (e.g. signs, other vehicles).  To create an adversarially attacked video from a clean video in BDD100K, an adversarial patch is added at a random time between 1 and 10 seconds into the video at a random position.  The patch is then applied to each following frame in the video, with the position changed using a random walk model in which the patch is moved at a random speed to random waypoints throughout the video.  Scale at time $t$ seconds is determined by a randomly generated sinusoidal function ranging between 0.2 and 2. This process moves the patch smoothly throughout the video like objects in the real world, and creates a diverse sampling of patch positions and scales.

\subsection{Evaluation Dataset}
A evaluation set was created by randomly selecting 100 videos from the BDD100K dataset.  Those 100 videos were added to the evaluation set as clean examples, and were then used to create 100 synthetic adversarial videos, for a evaluation set with 200 videos totaling to 240000 frames or 133 minutes.

\subsection{Attack Detection Rate}
ADAV's two-stage process requires it to accurately detect the presence of a full scale patch (detecting the presence of a smaller scale patch is less important since smaller patches create significantly weaker attacks).  Therefore, we measure Attack Detection Rate separately for scales greater than $0.8$.  Because ADAV either determines a frame to be clean or adversarially attacked, we can measure Attack Detection Rate with metrics for binary classification (Table~\ref{table1}).

\begin{table}
\caption{Attack Detection Rate}\label{table1}
\centering
\begin{tabular}{|c|c|c|c|}
  \hline
   &\textbf{Accuracy}&\textbf{Precision}&\textbf{Recall}\\ \hline
  \textbf{Patch Scale$>$0.8}&0.88&0.83&0.95\\ \hline
  \textbf{Patch Scale$<$0.8}&0.67&0.71&0.52\\ \hline
\end{tabular}
\end{table}

\subsection{Defense Performance}
The performance of the object detector after running the defense on both clean and adversarial inputs can be measured with standard metrics for evaluating object detectors, we choose mAP@IoU=50 (the threshold for a valid detection is 50\% intersection with the ground truth box).  

LGS, JPEG Compression, and Universal Defense Frames were used as baselines, as they are human-out-of-the-loop and can be computed in a realistic amount of time. 

Additionally, the frames per second (FPS) of each defense on a T4 GPU was measured to determine inference time latency.

The results of this evaluation are in Table~\ref{table2}.  The adversarial and clean performance of each defended model were measured.  Additionally, because ADAV processes clean and adversarial inputs differently, clean and adversarial latency were measured separately.  

\begin{table}
\caption{Defense Performance}\label{table2}
\centering
\begin{tabular}{|c|c|c|c|c|}
  \hline
  \textbf{Defense} & \textbf{Adversarial mAP} & \textbf{Clean mAP} & \textbf{Adversarial FPS} & \textbf{Clean FPS} \\ \hline
  No Defense & 0.22 & 0.46 & 63 & 63 \\ \hline
  LGS ($\lambda=2.3$) & 0.30 & 0.38 & 35 & 35 \\ \hline
  \makecell{JPEG Compression \\ (Quality=20\%)} & 0.23 & 0.41 & 8 & 8 \\ \hline
  \makecell{Universal \\ Defense Frame} & 0.20 & 0.43 & \textbf{62} & \textbf{63} \\ \hline
  ADAV & \textbf{0.36} & \textbf{0.44} & 20 & 56 \\ \hline
\end{tabular}
\end{table}

\subsection{Analysis}
It is critical for ADAV to flag all adversarially attacked frames so that the second stage can mask the patch.  However, it is less important to have a low false positive rate, since even if an attack is falsely detected, the second stage may not find any suspicious regions on a clean image, leaving the clean input unchanged.  ADAV aligns with these goals, since ADAV has a very high recall compared to precision, suggesting that ADAV is flagging almost all adversarially attacked frames while making some false positives.

Additionally, ADAV demonstrates high performance in defending against adversarial patch attacks by localizing and masking out patches.  ADAV significantly outperforms the next-best LGS in adversarial performance, and also exhibits higher clean performance due to its two-stage approach, which leaves clean inputs unchanged (unlike LGS).  ADAV also suffers almost no loss in FPS on clean videos, suggesting that ADAV can always be active in an AV's perception system.

\section{Discussion and Conclusion}
In this paper, we propose ADAV, a novel defense that specifically focuses on object detection in an AV context.  ADAV is designed with the unique characteristics of self-driving in mind, as it defends an object detection model trained on a driving dataset, takes advantage of the contextual information videos provide using temporal consistency, and is able to run in real-time.

Some improvements to ADAV could be using inpainting techniques to reduce information loss from regions obscured by patches, and using different attribution methods such as Guided GradCAM to produce less noisy saliency maps.

Regarding concerns of dynamically generated patches, such attacks are still bound by the temporal consistency check.  Other adversarial attacks such as adversarial perturbations are harder to execute against AVs in the real world because they are less robust to transformations such as lighting or camera noise, so they were not the focus of this work.  However, if temporal consistency is broken, ADAV's first stage can still effectively detect the presence of the attacks. Extending ADAV's second stage to address such attacks can be a direction for future work.

%
%
%
%
\bibliographystyle{splncs04}
\bibliography{bibtexes}
\end{document}